# Beyond Words: How Large Language Models Perform in Quantitative Management Problem-Solving

Jonathan Kuzmanko

# Abstract

This study examines how Large Language Models (LLMs) perform when tackling quantitative management decision problems in a zero-shot setting. Drawing on 900 responses generated by five leading models across 20 diverse managerial scenarios, our analysis explores whether these base models can deliver accurate numerical decisions under varying presentation formats, scenario complexities, and repeated attempts. Contrary to prior findings, we observed no significant effects of text presentation format (direct, narrative, or tabular) or text length on accuracy. However, scenario complexity—particularly in terms of constraints and irrelevant parameters—strongly influenced performance, often degrading accuracy. Surprisingly, the models handled tasks requiring multiple solution steps more effectively than expected. Notably, only 28.8% of responses were exactly correct, highlighting limitations in precision. We further found no significant "learning effect" across iterations: performance remained stable across repeated queries. Nonetheless, significant variations emerged among the five tested LLMs, with some showing superior binary accuracy. Overall, these findings underscore both the promise and the pitfalls of harnessing LLMs for complex quantitative decision-making, informing managers and researchers about optimal deployment strategies.

# Introduction

The integration of generative AI technologies such as Large Language Models (LLMs) into business environments represents a significant shift in organizational decision-making practices. The generative AI integration into business processes can enhance decision-making and strategic planning, aligning with the potential for operational efficiency improvements (Singh et al., 2024). This trend is particularly notable as AI adoption enhances augmented decision-making by strengthening team perceptions, improving assessment accuracy, and accelerating real-time decisions (Gama & Magistretti, 2023).

While LLMs demonstrate impressive capabilities in various domains, including achieving superior performance across numerous natural language processing tasks (Min et al., 2024), they also face significant challenges. Research shows that LLMs can produce hallucinations that lead to statements that seem plausible but are factually incorrect (Romera-Paredes et al., 2024). This limitation is particularly critical in business contexts where decision accuracy is paramount. Of particular concern is that neural network models, while showing strong mathematical reasoning capabilities, still have underdeveloped fundamental abilities in number representation and manipulation (Testolin, 2024). This raises important questions about their reliability in quantitative management decision tasks, especially when used basic models without task-specific training.

This research examines whether base LLMs, used in a zero-shot setting without specific examples or fine-tuning, can effectively assist in solving complex management problems characterized by "few-steps" quantitative calculations and inherent constraints. We investigate how factors such as information presentation complexity, solution steps, and irrelevant parameters affect the accuracy of LLMs in management decision-making tasks. By comparing five leading language models across multiple iterations and varying presentation formats, this study provides insights into the reliability and consistency of LLM-assisted decision-making in management contexts.

The research addresses three primary questions:

How does information presentation complexity affect accuracy of management quantitative tasks in LLMs?

How do scenario complexity factors influence output quantitative accuracy?

Do language models demonstrate performance differences across iterations and models?

Understanding these aspects is crucial as organizations increasingly integrate LLMs into their decision-making processes. The findings will contribute to both theoretical understanding of LLM capabilities and practical guidelines for managers seeking to leverage these tools effectively in quantitative decision-making tasks. xd

# 1  Theoretical Background

Recent advances in LLMs have transformed how organizations approach decision-making and problem-solving tasks. Research demonstrates significant progress in LLMs' capabilities across various domains, particularly in their ability to handle complex reasoning and decision tasks.

## 1.1      Current State of LLM Capabilities

Recent studies reveal that LLMs have achieved remarkable progress in handling complex tasks. As demonstrated in research by Ryu et al. (2024), GPT3.5 exhibits decision effects that parallel those observed in human subjects, suggesting these models can replicate human-like decision-making patterns. The advancement is particularly notable in business contexts, where with the advent of LLM-powered software, between 47 and 56% of all tasks could be completed significantly faster at the same level of quality (Eloundou et al., 2023).

## 1.2      Mathematical and Quantitative Reasoning

However, when it comes to mathematical and quantitative reasoning, research identifies specific limitations. While neural network models show strong mathematical reasoning capabilities, they still lack fundamental skills in number representation and manipulation (Testolin, 2024). This observation is particularly relevant for management decision-making, where quantitative accuracy is crucial.

Studies indicate that information presentation significantly impacts performance. Research by Yu et al. (2024) has shown that the efficiency of pre-trained LLMs in reasoning tasks depends heavily on how information is presented, with structured formats like tables yielding better performance than narrative or unstructured text. This finding suggests that the way quantitative information is presented could significantly affect decision accuracy.

Recent research has demonstrated that presentation format can significantly impact LLM performance, with variations of up to 40% based on the prompt structure used (He et al., 2024). Models show improved efficiency when working with logical and structured formats, achieving up to 5.7% better reasoning performance compared to natural language (Chen et al., 2024). However, text length appears to have a complex relationship with performance, as extending inputs beyond optimal length can degrade reasoning capabilities (Levy et al., 2024).

## 1.3 Complexity and Performance

Research shows that task complexity substantially influences LLM performance. As Zhang et al. (2024) demonstrated, tasks that require multiple transformations and higher reasoning steps result in decreased accuracy. This limitation becomes more pronounced when dealing with numerical computations, as generic neural networks, including advanced transformer-based architectures, face challenges with extrapolation and often fail when performing tasks that require intermediate calculations (Testolin, 2024).

## 1.4 Learning Effects and Model Behavior

An important aspect of LLM performance is their potential for learning and adaptation. According to Zhang et al. (2024), when question-answering involves intermediate data refinement, performance tends to improve across subsequent iterations. Research indicates that LLMs demonstrate some capacity for improvement through reinforcement learning and error analysis (Monea et al., 2024; Ying et al., 2024). However, this capability appears to be context-dependent, as LLMs excel at intuitive, stepwise tasks that use few examples or instructions, but face difficulties with tasks requiring multistep reasoning (Ryu et al., 2024).

## 1.5 Challenges and Limitations of LLMs

Several challenges remain in LLM implementation. As noted by Romera-Paredes et al. (2024), these models can generate hallucinations that lead to statements which appear plausible but are factually incorrect. Furthermore, when reasoning tasks contain more irrelevant or noisy parameters, they are more likely to produce suboptimal or inaccurate predictions (Yu et al., 2024). Comparative analyses have also revealed significant performance variations between LLM models, with different models showing distinct strengths in specific tasks (Haq et al., 2024; Akter et al., 2023)

## 1.6 Research Questions and Hypotheses

These findings lead us to several key research questions about LLMs' capabilities in quantitative management decision tasks:

**RQ1: How does information presentation complexity affect accuracy of management quantitative tasks in LLMs?**

H1a: Presentation format (direct/narrative/table) affects answer's accuracy

H1b: Text length of problem presentation negatively correlates with accuracy of model's answers

**RQ2: How do scenario complexity factors influence accuracy?**

H2a: Number of constraints negatively affects accuracy

H2b: Number of solution steps negatively affects accuracy

H2c: Number of irrelevant parameters negatively affects accuracy

**RQ3: Do LLMs demonstrate learning effects and performance differences?**

H3a: The models answers improved over iterations

H3b: There are significant differences between models' performance in both binary and distance from optimal answers.

**Figure 1**: *Research Model - Factors Influencing LLM Decision Accuracy in Quantitative Management decision tasks*

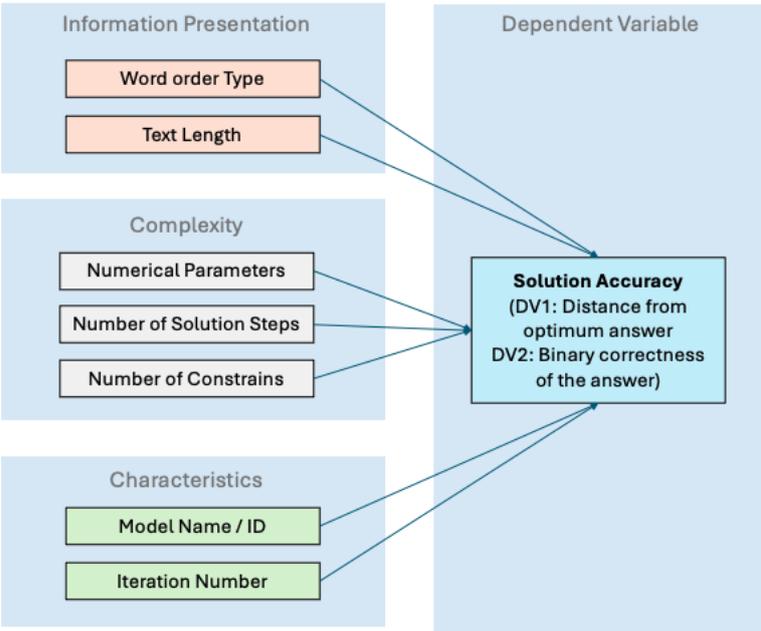

# 2 Research Methods

This study employed a systematic approach to evaluate LLMs' performance in quantitative management decision-making tasks. We developed and tested 20 management scenarios (a managerial question with an optimal numerical answer, such as optimal budget allocation, calculation of work hours required to complete a task, etc.). Each scenario presented in three different text formats and tested across three iterations answered by five LLM's. This design allowed us to examine how various factors, including information presentation, scenario complexity, and model characteristics, influence decision accuracy.

## 2.1 Sample

The dataset comprises 900 observations collected through systematic answers of five leading Large Language Models: Llama 3.3 70b, Gemeni 1.5 Pro, Grok, GPT4o, and Claude 3.5 Sonnet. These models were selected based on specific criteria including release date (post-2023), model size (minimum 30 billion parameters), context window capacity (minimum 8,000 tokens), user base size (>100,000 monthly active users), documented relevance to managerial decision-making in various domains. Together, these models represent the current "top tier" in publicly available large language models, offering a comprehensive view of current capabilities in AI-assisted decision making.

The models were tested on 20 quantitative management scenarios. Each scenario was presented in three different formats: direct presentation (average length 198 characters), narrative format (355 characters), and tabular format (227 characters). Every format variation was tested three times with each model, with iterations separated by 24-hour intervals, resulting in 900 total observations (20 scenarios × 3 formats × 3 iterations × 5 models). All observations were included in the analysis, with no exclusions.

Each scenario incorporated between 3-5 base constraints and required 2-5 solution steps (for reach the optimal answer), reflecting varying levels of complexity. The scenarios included both relevant parameters necessary for calculation and randomly distributed irrelevant parameters. All testing was conducted through the sdk.vercel.ai platform, ensuring standardized evaluation conditions across models. A consistent zero-shot prompting approach was employed, with models receiving identical instructions to act as management consultants and provide numerical solutions without explanations or additional context.

## 2.2 Measures and Data Collection

### 2.2.1 Dependent Variables

The study employed two primary measures of model answer accuracy: binary accuracy and deviation (percentage distance) from the optimal solution. Binary accuracy (Binary_Opt) was recorded to indicate exact matches of model's answer with the optimal numerical solution, providing a straightforward binary assessment of model correctness. The deviation from the optimal solution was measured as absolute logarithmic distance (log_distance), calculated as |ln(model_answer/optimal_solution)|. The logarithmic distance measure was chosen for its ability to normalize the effect of different numerical scales across business scenarios, treat proportional differences equally, and reduce the impact of extreme values while maintaining the relative importance of smaller deviations.

### 2.2.2 Independent Variables

The study examined three categories of independent variables. (1) Text characteristics included presentation of scenario format (direct, narrative, or tabular), text length of the prompt (measured in characters), number of numerical parameters (relevant and irrelevant) at the question. (2) The solution complexity, measured by the number of solution steps (the minimum required steps to reach the optimal solution) and constraints (the number of constrains in the question). (3) Model characteristics included model identity (name of model) and iteration number.

### 2.2.3 Data Collection Procedure

Data collection was conducted over a 15-day period through the *sdk.vercel.ai* platform using standardized one-shot prompting. Due to platform limitations, data collection was structured in batches of 60 questions per 24-hour period. Each question was asked in a new chat session to ensure independence of responses, with logs maintained through the *vercel.ai* platform for verification purposes.

The data collection process was designed to align with our research questions, particularly regarding response consistency across iterations. No responses were excluded from the analysis, and there were no missing values in the dataset. Quality control measures focused primarily on ensuring technical validity of responses, monitoring for potential technical interruptions or model failures. Given that response consistency was one of our primary research interests, we deliberately did not implement consistency checks or corrections during data collection, allowing natural variations in model responses to be captured and analyzed.

#### 2.2.4 Measurement Limitations

Our measurement approach faces two main technical limitations. First, while Vercel is a reliable and widely accepted development platform, the use of API access rather than direct model interaction represents a potential technical constraint. Second, the focus on measuring distance from optimal solutions, while providing granular performance insights, may not fully reflect real-world decision-making contexts where binary outcomes might be more relevant. Additionally, the standardized zero-shot prompting approach, while necessary for experimental control, may not capture all real-world usage scenarios.

## 2.3 Analysis

To test our research hypotheses, we employed multiple statistical approaches using Jamovi statistical software. Initial analysis included descriptive statistics and tests of statistical assumptions. We conducted Shapiro-Wilk tests to assess normality of distributions for all variables, complemented by examination of skewness and kurtosis. Due to identified non-normal distributions (all Shapiro-Wilk tests $p < .001$), we proceeded with non-parametric analyses where appropriate.

Due to non-normal distribution of our dependent variables, we combined logistic regression with non-parametric tests.

For examining scenario complexity effects (H1a-H1b), we specified a comprehensive logistic regression model for binary accuracy and Spearman correlations for log_distance:

$\text{logit}(\text{Binary\_Opt}) = \beta_0 + \beta_1 \text{Irrelevant\_Params\_Count} + \beta_2 \text{Solution\_Steps} + \beta_3 \text{Base\_Constraints\_Count} + \beta_4 \text{Text\_Length} + \beta_5 \text{Word\_Order\_Type} + \beta_6 \text{Model\_Name} + \beta_7 \text{Iteration\_Number} + \varepsilon$

For analyzing text presentation influence (H2a-H2b), we employed both the logistic regression model above and Kruskal-Wallis tests to examine the relationship between presentation formats and log_distance.

For evaluating learning effects and model differences (H3a-H3b), we utilized the logistic regression model for binary accuracy while employing Friedman tests to examine iteration effects on log_distance. Model differences in log_distance was assessed using Kruskal-Wallis tests.

We conducted preliminary tests to evaluate statistical assumptions. For logistic regression, these included multicollinearity assessment through VIF analysis to examine relationships between independent variables. Post-hoc comparisons using Dwass-Steel-Critchlow-Fligner test were conducted for significant Kruskal-Wallis results to examine specific group differences, chosen for its robust control of Type I error rate with equal group sizes.

# 3 Results

## 3.1 General Descriptive Statistics

The analysis included 900 observations across five LLMs, testing their performance in quantitative management decision tasks. Two primary accuracy measures were employed: a continuous measure (logarithmic distance from optimal solution) and a binary measure (correct/incorrect solution).

**Table 1**: *Descriptive Statistics of Primary Variables*

Descriptives

| | log_distance | Text_Length | Irrelevant_Params_Count | Solution_Steps | Base_Constraints_Count | Binary_Opt (1=yes, 0=no) |
|---|---|---|---|---|---|---|
| N | 900 | 900 | 900 | 900 | 900 | 900 |
| Mean | 0.252 | 260 | 8.72 | 3.05 | 3.75 | 0.288 |
| Median | 0.163 | 225 | 9.00 | 3.00 | 4.00 | 0.00 |
| Sum | 227 | 234075 | 7845 | 2745 | 3375 | 259 |
| Standard deviation | 0.318 | 80.8 | 3.84 | 0.921 | 0.699 | 0.453 |
| Minimum | 0.00 | 142 | 2 | 2 | 3 | 0 |
| Maximum | 2.40 | 465 | 18 | 5 | 5 | 1 |
| Skewness | 2.74 | 0.677 | 0.667 | 0.671 | 0.386 | 0.939 |
| Std. error skewness | 0.0815 | 0.0815 | 0.0815 | 0.0815 | 0.0815 | 0.0815 |
| Kurtosis | 11.8 | -0.614 | 0.412 | -0.304 | -0.918 | -1.12 |
| Std. error kurtosis | 0.163 | 0.163 | 0.163 | 0.163 | 0.163 | 0.163 |
| Shapiro-Wilk W | 0.737 | 0.909 | 0.928 | 0.834 | 0.787 | 0.567 |
| Shapiro-Wilk p | < .001 | < .001 | < .001 | < .001 | < .001 | < .001 |

The logarithmic distance from optimal solutions (log_distance) showed considerable variation ($M = 0.252$, $SD = 0.318$, $Mdn = 0.163$), with values ranging from 0 to 2.40. The binary accuracy measure (Binary_Opt) indicated that 28.8% of solutions were exactly correct. Text prompts varied substantially in length ($M = 260$ characters, $SD = 80.8$), ranging from 142 to 465 characters. The complexity parameters showed that scenarios typically contained multiple constraints ($M = 3.75$, $SD = 0.699$) and solution steps ($M = 3.05$, $SD = 0.921$).

Distribution analysis revealed non-normal distributions for all primary variables (Shapiro-Wilk tests, all p < .001). The log_distance measure showed particular right-skew (skewness = 2.74) and leptokurtosis (kurtosis = 11.8), indicating a concentration of responses near the optimal solution with some notable outliers.

**Figure 2**: *Distribution of Logarithmic Distance from Optimal Solutions*

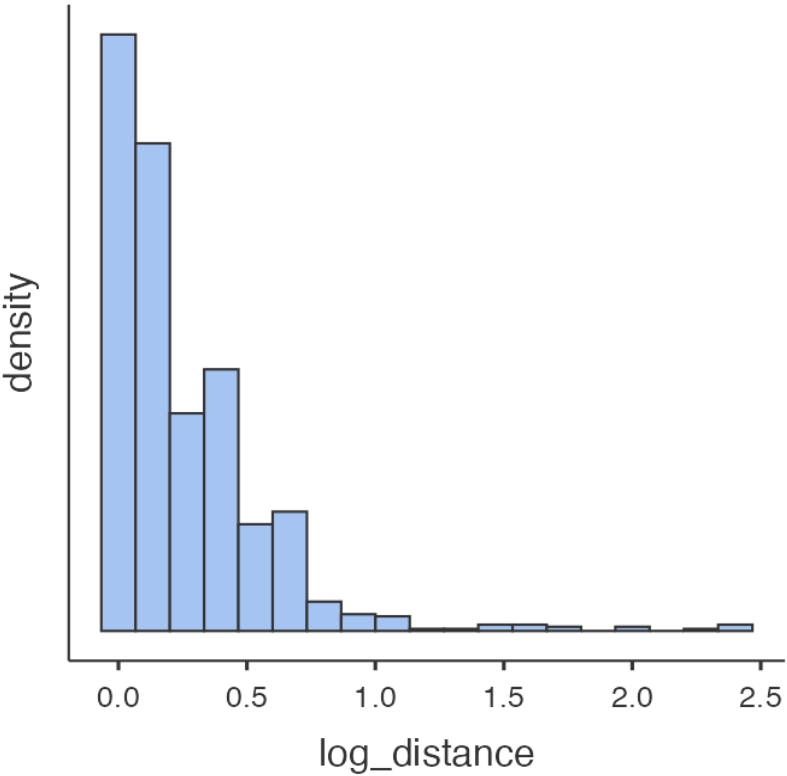

The histogram and illustrates the substantial right-skew in logarithmic distance scores, with most responses clustered near optimal solutions but a notable right tail of larger deviations. This distribution pattern may be attributed to our calculation method for the distance variable, which used absolute values. Nevertheless, this computational approach does not affect the fundamental non-normality of the variable.

**Table 2**: *Inter-correlation Matrix of Study Variables*

Correlation Matrix

| | | log_distance | Binary_Opt (1=yes, 2=no) | Text_Length | Irrelevant_Params_Count | Numerical_Params_Count | Solution_Steps | Base_Constraints_Count | Iteration_Number |
|---|---|---|---|---|---|---|---|---|---|
| log_distance | Spearman's rho | — | | | | | | | |
| | df | — | | | | | | | |
| | p-value | — | | | | | | | |
| Binary_Opt (1=yes, 2=no) | Spearman's rho | −0.794 *** | — | | | | | | |
| | df | 898 | — | | | | | | |
| | p-value | <.001 | — | | | | | | |
| Text_Length | Spearman's rho | 0.014 | −0.027 | — | | | | | |
| | df | 898 | 898 | — | | | | | |
| | p-value | 0.672 | 0.414 | — | | | | | |
| Irrelevant_Params_Count | Spearman's rho | 0.039 | −0.034 | 0.480 *** | — | | | | |
| | df | 898 | 898 | 898 | — | | | | |
| | p-value | 0.243 | 0.302 | <.001 | — | | | | |
| Numerical_Params_Count | Spearman's rho | 0.130 *** | −0.109 ** | 0.518 *** | 0.969 *** | — | | | |
| | df | 898 | 898 | 898 | 898 | — | | | |
| | p-value | <.001 | 0.001 | <.001 | <.001 | — | | | |
| Solution_Steps | Spearman's rho | 0.126 *** | −0.103 ** | 0.160 *** | 0.097 ** | 0.199 *** | — | | |
| | df | 898 | 898 | 898 | 898 | 898 | — | | |
| | p-value | <.001 | 0.002 | <.001 | 0.004 | <.001 | — | | |
| Base_Constraints_Count | Spearman's rho | 0.294 *** | −0.217 *** | 0.174 *** | 0.078 * | 0.233 *** | 0.353 *** | — | |
| | df | 898 | 898 | 898 | 898 | 898 | 898 | — | |
| | p-value | <.001 | <.001 | <.001 | 0.020 | <.001 | <.001 | — | |
| Iteration_Number | Spearman's rho | −0.023 | 0.039 | 0.000 | 0.000 | 0.000 | 0.000 | 0.000 | — |
| | df | 898 | 898 | 898 | 898 | 898 | 898 | 898 | — |
| | p-value | 0.487 | 0.242 | 1.000 | 1.000 | 1.000 | 1.000 | 1.000 | — |

*Note.* * p < .05, ** p < .01, *** p < .001

Correlation analysis revealed several significant relationships between performance measures and task characteristics. The logarithmic distance showed moderate positive correlations with the number of base constraints (rs = .294, p < .001) and weaker but significant correlations with solution steps (rs = .126, p < .001) and numerical parameters (rs = .130, p < .001). Text length showed no significant correlation with accuracy measures (rs = .014, p = .672).

The non-normal distributions and presence of significant correlations between task characteristics influenced our choice of statistical methods, leading to the use of non-parametric tests for many subsequent analyses.

Prior to hypothesis testing, we examined multicollinearity between predictors. VIF analysis showed acceptable levels for most relevant variables (VIF < 5).

**Table 3**: *Multicollinearity Assessment of Predictor Variables*

|                         | VIF   | Tolerance |
|-------------------------|-------|-----------|
| Text_Length             | 2.492 | 0.401     |
| Numerical_Params_Count  | 7.647 | 0.131     |
| Base_Constraints_Count  | 1.317 | 0.759     |
| Solution_Steps          | 1.295 | 0.772     |
| Irrelevant_Params_Count | 7.453 | 0.134     |
| Word_Order_Type         | 1.597 | 0.626     |
| Model_Name              | 1.000 | 1.000     |
| Iteration_Number        | 1.000 | 1.000     |

## 3.2 Primary Findings

Analysis of LLM performance in quantitative decision-making tasks revealed distinct patterns across models, presentation formats, and complexity levels.

### 3.2.1 Model Performance Differences

**Table 4**: *Descriptive Statistics by Model Type*

Descriptives

|  | Model_Name | N | Mean | Median | SD |
|---|---|---|---|---|---|
| log_distance | claude-3.5-sonnet | 180 | 0.205 | 0.129 | 0.251 |
|  | gemeni-1.5-pro | 180 | 0.286 | 0.154 | 0.343 |
|  | gpt-4o | 180 | 0.228 | 0.148 | 0.264 |
|  | grok | 180 | 0.246 | 0.163 | 0.347 |
|  | llama-3.3-70b | 180 | 0.295 | 0.179 | 0.361 |

Claude-3.5-sonnet demonstrated the highest overall accuracy (M = 0.205, SD = 0.251), followed by GPT-4o (M = 0.228, SD = 0.264). Llama-3.3-70b and Gemeni-1.5-pro showed relatively higher logarithmic distances (M = 0.295, SD = 0.361 and M = 0.286, SD = 0.343 respectively), indicating lower accuracy.

**Figure 3**: *Model Performance Distribution*

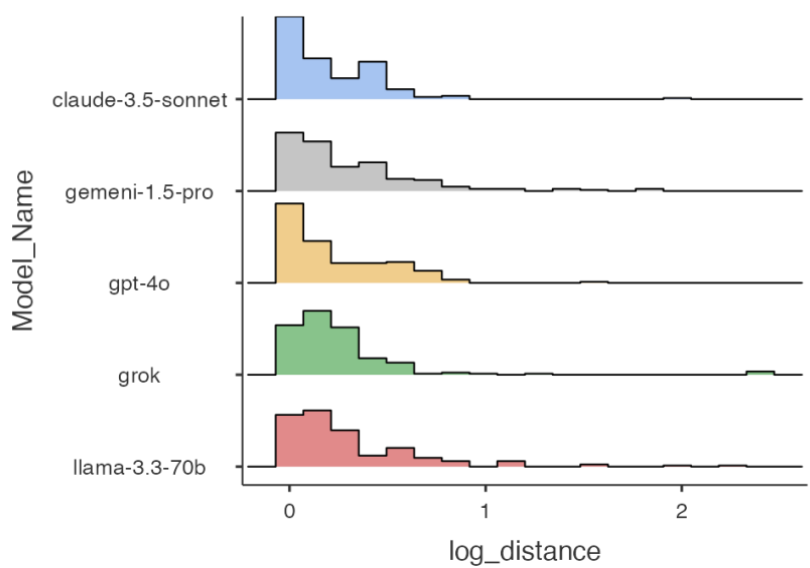

**Figure 4**: *Model Performance Box Plots*

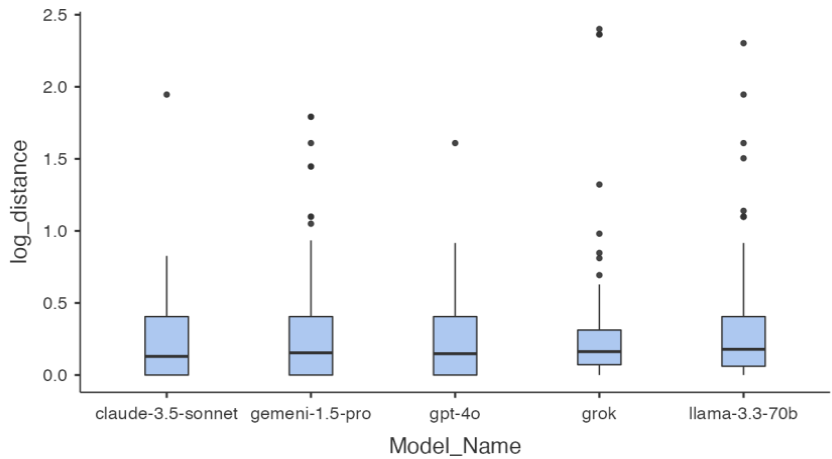

Differences in performance were tested using the Kruskal-Wallis test. For log distance, no significant differences were found ($\chi^2 = 8.52$, df = 4, p = 0.074, $\varepsilon^2 = 0.00947$). For binary accuracy, stronger difference was observed between models ($\chi^2 = 14.15$, df = 4, p = 0.007, $\varepsilon^2 = 0.01575$).

**Table 5**: *Non-parametric Analysis of Model Differences*

Kruskal-Wallis

|  | $\chi^2$ | df | p | $\varepsilon^2$ |
|---|---|---|---|---|
| log_distance | 8.52 | 4 | 0.074 | 0.00947 |
| Binary_Opt (1=yes, 0=no) | 14.15 | 4 | 0.007 | 0.01575 |

Post-hoc comparisons revealed significant differences between specific model pairs:

**Table 6**: *Pairwise Performance Comparisons*

Pairwise comparisons - log_distance

| | | W | p |
|---|---|---|---|
| claude-3.5-sonnet | gemeni-1.5-pro | 2.995 | 0.213 |
| claude-3.5-sonnet | gpt-4o | 1.071 | 0.943 |
| claude-3.5-sonnet | grok | 2.013 | 0.613 |
| claude-3.5-sonnet | llama-3.3-70b | 3.722 | 0.065 |
| gemeni-1.5-pro | gpt-4o | -1.937 | 0.647 |
| gemeni-1.5-pro | grok | -0.944 | 0.963 |
| gemeni-1.5-pro | llama-3.3-70b | 0.529 | 0.996 |
| gpt-4o | grok | 1.065 | 0.944 |
| gpt-4o | llama-3.3-70b | 2.676 | 0.322 |

Pairwise comparisons - log_distance

|  |  | W | p |
|---|---|---|---|
| grok | llama-3.3-70b | 1.234 | 0.907 |

**Table 7**: *Binary Accuracy Comparisons*

Pairwise comparisons - Binary_Opt (1=yes, 0=no)

|  |  | W | p |
|---|---|---|---|
| claude-3.5-sonnet | gemeni-1.5-pro | -2.867 | 0.253 |
| claude-3.5-sonnet | gpt-4o | -1.090 | 0.939 |
| claude-3.5-sonnet | grok | -3.878 | 0.048 |
| claude-3.5-sonnet | llama-3.3-70b | -4.397 | 0.016 |
| gemeni-1.5-pro | gpt-4o | 1.782 | 0.716 |
| gemeni-1.5-pro | grok | -1.024 | 0.951 |
| gemeni-1.5-pro | llama-3.3-70b | -1.553 | 0.808 |
| gpt-4o | grok | -2.800 | 0.276 |
| gpt-4o | llama-3.3-70b | -3.324 | 0.129 |
| grok | llama-3.3-70b | -0.530 | 0.996 |

For log_distance, pairwise comparisons (table 6) did not reveal statistically significant differences between most models, though the comparison between Claude-3.5-sonnet and Llama-3.3-70b approached significance (W = 3.722, p = 0.065). Pairwise comparisons for binary accuracy (table 7) revealed near to significant differences between Claude-3.5-sonnet and both Grok (W = -3.878, p = 0.048) and Llama-3.3-70b (W = -4.397, p = 0.016).

### 3.2.2 Complexity Effects

Different aspects of task complexity showed varying impacts on model performance. Kruskal-Wallis analysis revealed significant effects of task complexity components:

**Table 8**: *Non-parametric Analysis of Solution Steps*

Kruskal-Wallis

|  | $\chi^2$ | df | p | $\varepsilon^2$ |
|---|---|---|---|---|
| log_distance | 43.2 | 3 | < .001 | 0.0480 |
| Binary_Opt (1=yes, 0=no) | 29.2 | 3 | < .001 | 0.0325 |

Analysis revealed significant effects of solution steps on both logarithmic distance ($\chi^2$ = 43.2, df = 3, p < .001, $\varepsilon^2$ = 0.0480) and binary accuracy ($\chi^2$ = 29.2, df = 3, p < .001, $\varepsilon^2$ = 0.0325).

**Table 9:** *Solution Steps Pairwise Comparisons*

Pairwise comparisons - log_distance

|   |   | W | p |
|---|---|---|---|
| 2 | 3 | 5.1995 | 0.001 |
| 2 | 4 | 8.5055 | < .001 |
| 2 | 5 | 0.0391 | 1.000 |
| 3 | 4 | 5.0337 | 0.002 |
| 3 | 5 | -3.7857 | 0.037 |
| 4 | 5 | -6.4423 | < .001 |

Dwass-Steel-Critchlow-Fligner pairwise comparisons revealed a consistent pattern of increasing complexity effects. Significant differences were found between two-step problems and both three-step (W = 5.1995, p = 0.001) and four-step problems (W = 8.5055, p < .001). Additionally, four-step problems showed significantly higher complexity effects compared to three-step problems (W = 5.0337, p = 0.002) and five-step problems (W = -6.4423, p < .001), could indicate a non-linear complexity threshold effect peaking at four steps.

**Table 10**: *Complexity Components Analysis (constrains count)*

Kruskal-Wallis

|   | $\chi^2$ | df | p | $\varepsilon^2$ |
|---|---|---|---|---|
| log_distance | 96.9 | 2 | < .001 | 0.1078 |
| Binary_Opt (1=yes, 0=no) | 66.1 | 2 | < .001 | 0.0735 |

The impact of complexity was particularly pronounced when examining multiple components simultaneously, with significant effects observed for both logarithmic distance ($\chi^2$ = 96.9, p < .001, $\varepsilon^2$ = 0.1078) and binary accuracy ($\chi^2$ = 66.1, p < .001, $\varepsilon^2$ = 0.0735).

### 3.2.3 Presentation Format Patterns

**Table 11**: *Performance Across Presentation Formats*

Descriptives

|  | Word_Order_Type | log_distance | Text_Length |
|---|---|---|---|
| N | Direct | 300 | 300 |
|  | Narrative | 300 | 300 |
|  | Table | 300 | 300 |
| Mean | Direct | 0.225 | 198 |
|  | Narrative | 0.226 | 355 |
|  | Table | 0.305 | 227 |
| Median | Direct | 0.154 | 188 |
|  | Narrative | 0.154 | 345 |
|  | Table | 0.182 | 225 |
| Standard deviation | Direct | 0.272 | 49.5 |
|  | Narrative | 0.272 | 46.1 |
|  | Table | 0.389 | 30.8 |

Different presentation formats showed varying effects on model performance. Initial analysis indicated that tabular presentations showed higher logarithmic distances (M = 0.305, SD = 0.389) compared to direct (M = 0.225, SD = 0.272) and narrative formats (M = 0.226, SD = 0.272).

**Table 12**: *Non-parametric Analysis of Presentation Format (word type)*

Kruskal-Wallis

|  | $\chi^2$ | df | p | $\varepsilon^2$ |
|---|---|---|---|---|
| log_distance | 5.124 | 2 | 0.077 | 0.00570 |
| Binary_Opt (1=yes, 0=no) | 0.856 | 2 | 0.652 | 9.52e-4 |

However, these differences did not reach statistical significance for either logarithmic distance ($\chi^2$ = 5.124, df = 2, p = 0.077, $\varepsilon^2$ = 0.00570) or binary accuracy ($\chi^2$ = 0.856, df = 2, p = 0.652, $\varepsilon^2$ = 0.000952).

**Figure 5**: *Distribution of Performance by Presentation Format*

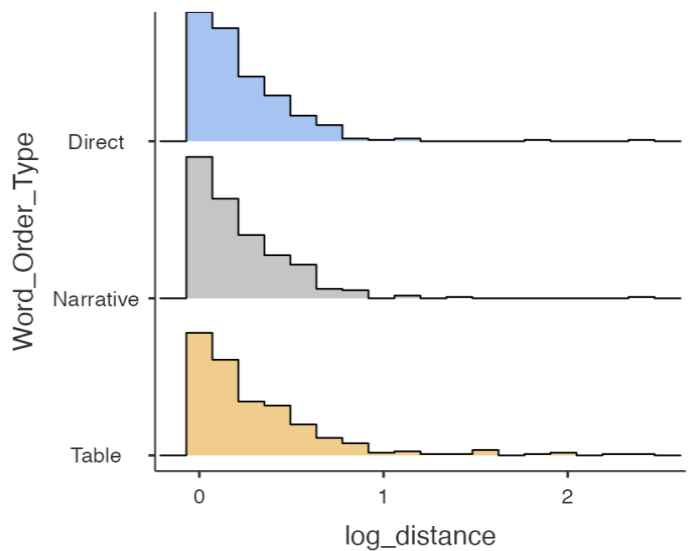

The distribution analysis reveals interesting patterns in how models process different presentation formats, though these effects were more subtle than the complexity impacts. These format effects will be examined in detail in the hypothesis testing section.

## 3.3 Hypothesis Testing

### 3.3.1 Information Presentation Effects (H1a)

The first set of hypotheses examined the impact of presentation format on model accuracy. As shown in Table 8, initial descriptive statistics suggested potential differences between presentation formats. Further analysis using Kruskal-Wallis tests (Table 9) revealed that these differences did not reach statistical significance for either logarithmic distance or binary accuracy.

**Table 13**: *Overall Logistic Regression Model Fit (Model Fit Statistics for Binary Accuracy Prediction)*

Model Fit Measures

| Model | Deviance | AIC | BIC | $R^2_{McF}$ | $R^2_N$ |
|---|---|---|---|---|---|
| 1 | 945 | 971 | 1033 | 0.125 | 0.200 |

**Table 14**: *Omnibus Effects in Logistic Regression*

Omnibus Likelihood Ratio Tests

| Predictor | χ² | df | p |
|---|---|---|---|
| Text_Length | 0.00153 | 1 | 0.969 |
| Numerical_Params_Count | 80.08868 | 1 | < .001 |
| Base_Constraints_Count | 0.54709 | 1 | 0.460 |
| Iteration_Number | 1.62484 | 1 | 0.202 |
| Irrelevant_Params_Count | 77.48279 | 1 | < .001 |
| Word_Order_Type | 0.60576 | 2 | 0.739 |
| Model_Name | 16.39910 | 4 | 0.003 |
| Solution_Steps | 13.52069 | 1 | < .001 |

**Table 15**: *Logistic Regression Coefficients*

Model Coefficients - Binary_Opt (1=yes, 0=no)

| Predictor | Estimate | 95% Confidence Interval Lower | 95% Confidence Interval Upper | SE | Z | p |
|---|---|---|---|---|---|---|
| Intercept | 5.4260 | 4.03875 | 6.81325 | 0.70779 | 7.6661 | <.001 |
| Text_Length | −1.01e−4 | −0.00514 | 0.00494 | 0.00257 | −0.0391 | 0.969 |
| Numerical_Params_Count | −1.2818 | −1.57622 | −0.98729 | 0.15024 | −8.5313 | <.001 |
| Base_Constraints_Count | −0.1065 | −0.38830 | 0.17528 | 0.14377 | −0.7408 | 0.459 |
| Iteration_Number | 0.1251 | −0.06748 | 0.31766 | 0.09825 | 1.2731 | 0.203 |
| Irrelevant_Params_Count | 1.2982 | 0.99633 | 1.60011 | 0.15403 | 8.4284 | <.001 |
| Word_Order_Type: | | | | | | |
|   Narrative – Direct | 0.0999 | −0.76048 | 0.96032 | 0.43899 | 0.2276 | 0.820 |
|   Table – Direct | −0.1287 | −0.58755 | 0.33012 | 0.23410 | −0.5498 | 0.582 |
| Model_Name: | | | | | | |
|   gemeni-1.5-pro – claude-3.5-sonnet | −0.5371 | −1.01944 | −0.05475 | 0.24610 | −2.1824 | 0.029 |
|   gpt-4o – claude-3.5-sonnet | −0.1984 | −0.66538 | 0.26868 | 0.23828 | −0.8324 | 0.405 |
|   grok – claude-3.5-sonnet | −0.7418 | −1.23633 | −0.24735 | 0.25230 | −2.9404 | 0.003 |
|   llama-3.3-70b – claude-3.5-sonnet | −0.8514 | −1.35333 | −0.34944 | 0.25610 | −3.3244 | <.001 |
| Solution_Steps | 0.4246 | 0.19756 | 0.65159 | 0.11583 | 3.6656 | <.001 |

The logistic regression analysis provided additional insights, controlling for other variables. The results showed that compared to direct format, neither narrative ($\beta = 0.0999$, SE = 0.43899, p = 0.820) nor tabular formats ($\beta = -0.1287$, SE = 0.23410, p = 0.582) significantly affected binary accuracy. This held true even when controlling for text length ($\beta = -1.01\mathrm{e}{-4}$, SE = 0.00257, p = 0.969) and other complexity factors.

These findings lead to the **not-supporting of both H1a**, suggesting that LLMs demonstrate similar performance across different presentation formats in quantitative management decision tasks without significant difference between the type of text presentation.

**H1b**, examining the relationship between text length and accuracy, was **also not supported**. As shown in Table 2 (correlation matrix), text length showed no significant correlation with either logarithmic distance (rs = 0.014, p = 0.672) or binary accuracy. The logistic regression results further confirmed this finding, with text length showing no significant effect on model performance when controlling for other factors.

These results collectively suggest that LLMs' performance in quantitative management decisions is relatively robust to variations in information presentation, contradicting initial expectations about format and length effects.

### 3.3.2 Scenario Complexity Effects (H2a, H2b, H2c)

The second set of hypotheses examined how different aspects of scenario complexity affected model performance. The analysis considered base constraints (H2a), solution steps (H2b), and irrelevant parameters (H2c).

The Kruskal-Wallis Analysis of Task Complexity Components (Table 8) revealed significant effects of solution steps on both logarithmic distance ($\chi^2 = 43.2$, df = 3, p < .001, $\varepsilon^2 = 0.0480$) and binary accuracy ($\chi^2 = 29.2$, df = 3, p < .001, $\varepsilon^2 = 0.0325$).

The logistic regression analysis (Tables 13-15) provided additional insights while controlling for other variables. Solution steps showed a significant positive effect on binary accuracy ($\beta = 0.4246$, SE = 0.11583, p < .001), contrary to the hypothesized negative relationship. The number of irrelevant parameters demonstrated the expected negative effect ($\beta = 1.2982$, SE = 0.15403, p < .001), strongly supporting H2c.

As shown in Table 2, Spearman's correlation analysis further supported these relationships, revealing significant associations between complexity measures and performance. Base constraints showed moderate correlations with both accuracy measures (rs = 0.294, p < .001 for logarithmic distance), while solution steps (rs = 0.126, p < .001) and numerical parameters (rs = 0.130, p < .001) showed weaker but significant correlations.

The results provided mixed support for the complexity hypotheses. While **H2a and H2c were supported**, demonstrating significant effects of base constraints and irrelevant parameters on model performance, **H2b was not supported** due to an unexpected positive relationship between solution steps and accuracy. These findings suggest that the relationship between task complexity and LLM performance is more nuanced than initially hypothesized.

### 3.3.3 Learning Effects and Model Differences (H3a, H3b)

The final set of hypotheses examined potential learning effects across iterations (H3a) and performance differences between models (H3b).

**Table 16**: *Descriptive Statistics by Iteration*

Descriptives

| | Iteration_Number | N | Mean | Median | SD | Shapiro-Wilk W | p |
|---|---|---|---|---|---|---|---|
| log_distance | 1 | 300 | 0.256 | 0.156 | 0.301 | 0.791 | < .001 |
| | 2 | 300 | 0.256 | 0.163 | 0.343 | 0.690 | < .001 |
| | 3 | 300 | 0.244 | 0.161 | 0.309 | 0.734 | < .001 |

Analysis of iteration effects revealed relatively stable performance across attempts (Iteration 1: M = 0.256, SD = 0.301; Iteration 2: M = 0.256, SD = 0.343; Iteration 3: M = 0.244, SD = 0.309).

**Table 17**: *Friedman Analysis of Log Distance Learning Effects* (*Non-parametric Analysis of Changes in Logarithmic Distance Across Iterations*)

Friedman

| $\chi^2$ | df | p |
|---|---|---|
| 0.0208 | 2 | 0.990 |

**Table 18**: *Friedman Analysis of Binary Accuracy Learning Effects* (*Non-parametric Analysis of Changes in Binary Accuracy Across Iterations*)

Friedman

| $\chi^2$ | df | p |
|---|---|---|
| 4.63 | 2 | 0.099 |

**Table 19**: *Non-parametric Analysis of iteration*

Kruskal-Wallis

|  | χ² | df | p | ε² |
|---|---|---|---|---|
| log_distance | 0.485 | 2 | 0.785 | 5.40e-4 |
| Binary_Opt (1=yes, 2=no) | 1.830 | 2 | 0.400 | 0.00204 |

The Friedman test for repeated measures showed no significant improvement of logd_distance across iterations ($\chi^2 = 0.0208$, df = 2, p = 0.990). The improve of binary correctness much stronger but still didn't reach the significant level ($\chi^2 = 4.63$, df = 2, p = 0.099). This finding was supported by Kruskal-Wallis analysis of both logarithmic distance ($\chi^2 = 0.485$, df = 2, p = 0.785, $\varepsilon^2 = 0.00054$) and binary accuracy ($\chi^2 = 1.830$, df = 2, p = 0.400, $\varepsilon^2 = 0.00204$).

These findings were further supported by the logistic regression analysis (Tables 13-15). The regression model showed no significant effect of iteration number on binary accuracy ($\beta = 0.1251$, p = 0.203), confirming the absence of learning effects. Model differences remained significant when controlling for other variables, with all models showing lower performance compared to Claude-3.5-sonnet: Gemeni-1.5-pro ($\beta = -0.5371$, p = 0.029), Grok ($\beta = -0.7418$, p = 0.003), and Llama-3.3-70b ($\beta = -0.8514$, p < .001)

Regarding model differences (H3b), as previously shown in Tables 4-7, significant variations in performance were observed between models. The Kruskal-Wallis test revealed significant differences in binary accuracy ($\chi^2 = 14.15$, df = 4, p = 0.007, $\varepsilon^2 = 0.01575$). Pairwise comparisons indicated that Claude-3.5-sonnet outperformed Grok (W = -3.878, p = 0.048) and Llama-3.3-70b (W = -4.397, p = 0.016) in binary accuracy.

Therefore, while **H3a was not supported**, showing no significant learning effects across iterations, **H3b was supported**, revealing performance differences between models in quantitative management decision tasks.

# 4 Discussion

## 4.1 Summary of Findings and Hypotheses Support

Our research examined the capabilities of LLMs in quantitative management decision tasks, focusing on three key aspects: information presentation, scenario complexity, and learning effects. The results provide several interesting insights, some of which challenge existing literature while others strengthen previous findings.

Regarding information presentation (RQ1), contrary to previous research suggesting that LLM reasoning efficiency is substantially affected by how information is presented (Yu et al., 2024), our findings showed no significant effect of presentation format on either binary accuracy or logarithmic distance from optimal solutions. This unexpected result might be attributed to the rapid change of LLM capabilities, as recent research indicates that newer models demonstrate increased robustness across different presentation formats (He et al., 2024). Regarding text length (H1b), while the existing literature shows mixed findings - with some studies finding that extended inputs can degrade performance (Levy et al., 2024) and others demonstrating improved accuracy with constrained length (Nayab et al., 2024) - our research found no significant correlation between text length and model performance.

The complexity hypotheses (RQ2) yielded mixed results. While H2a and H2c were supported, demonstrating the negative impact of base constraints and irrelevant parameters on accuracy, H2b revealed an unexpected positive relationship between solution steps and accuracy. This finding contrasts with Zhang et al.'s (2024) observation of diminishing accuracy with increased steps, suggesting that contemporary LLMs might have developed better capabilities in handling multi-step reasoning tasks. This evolution aligns with recent research showing that newer LLM versions are becoming more adept at handling complex reasoning chains (Akter et al., 2023; Chen et al., 2024; Zhang et al., 2024). However, the observed positive correlation between solution steps and accuracy is not fully understood and requires further investigation to elucidate the underlying mechanisms.

Regarding learning effects and model differences (RQ3), our findings partially aligned with existing literature. The absence of significant learning effects across iterations (H3a not supported) aligns with recent research highlighting that while LLMs exhibit human-like decision-making patterns, they fail to show improvement over repeated trials (Ryu et al., 2024). However, we found significant performance differences between different models (H3b supported), consistent with current literature highlighting varying capabilities across different LLMs (Haq et al., 2024; Akter et al., 2023; Wojcik et al., 2024). These differences often manifest in task-specific ways, with different models showing distinct strengths in particular types of problems.

A notable finding of this study, although not its primary focus, was the relatively low percentage of exactly correct answers (28.8%) achieved by the models in binary evaluation. This suggests that LLMs may face significant challenges in producing exact solutions to quantitative management problems. However, when evaluated on the continuous measure of logarithmic distance (log_distance), the models demonstrated more nuanced and clear performance, with an average distance of 0.252 (SD = 0.318).

The modest binary success rate and the variability in log_distance can be attributed to the constraints of our research design, including the use of zero-shot prompting without examples, limited variation in task parameters, and a relatively small set of base scenarios (20). While advanced prompting techniques or task-specific fine-tuning might have improved performance, these were deliberately excluded to maintain consistency with our focus on format effects, complexity impacts, and inter-model differences.

## 4.2  Validity Considerations and Limitations

While our dataset comprises 900 observations, it is important to note that these are derived from 20 base questions, each presented in different formats (word_type variable) and tested across multiple models. This structure, while providing robust testing for format and model effects, creates inherent limitations in the range and distribution of independent variables such as constraint counts and solution steps. The relatively small number of base scenarios constrains the possible combinations and ranges of these variables, potentially affecting our ability to fully capture their relationships with model performance. A study incorporating a wider variety of base scenarios would allow for broader ranges and more varied combinations of these independent variables, potentially providing additional insights into their effects on LLM performance.

The rapid evolution of LLM capabilities presents both a challenge and an opportunity for interpretation. Recent research has shown that "the performance of LLMs can vary dramatically based on the prompt format used" (He et al., 2024), suggesting that our findings might be sensitive to the specific prompting approaches employed (we have used basic zero-shot prompt).

Our measurement approach, particularly the use of logarithmic distance as a continuous measure of accuracy, provides a more nuanced view of model performance compared to binary accuracy alone. However, this approach may not capture all aspects of decision quality, particularly in cases where approximate solutions might be just practically acceptable in real-world contexts.

The generalizability of our findings should be considered within the context of quantitative management decision tasks. While our results provide insights into LLM capabilities in structured decision scenarios, the broader applicability to more complex, context-dependent management decisions may require additional investigation.

## 4.3 Practical Implications and Future Research Directions

The findings of this study have several important implications for the implementation of LLMs in management decision-making contexts. The observed differences in model performance suggest that organizations should implement systematic evaluation processes when selecting models for specific applications. While LLMs demonstrate significant potential for augmenting decision-making processes, their effective deployment requires careful consideration of both model capabilities and task characteristics.

The relationship between task complexity and model performance revealed in this study challenges some previous assumptions about LLM capabilities. The positive correlation between solution steps and accuracy, while unexpected, suggests that contemporary LLMs may be more adept at handling complex, multi-step problems than previously theorized. However, the negative impact of irrelevant parameters and base constraints emphasizes the importance of structured information presentation and careful prompt engineering in practical applications.

Future research should address current limitations through expanded scenario ranges to enable a more comprehensive analysis of task characteristics and model performance relationships. Additionally, given the rapid advancement in LLM technology, longitudinal studies examining performance changes across model versions could provide valuable insights into capability development patterns. Moreover, in light of the models' moderate success in delivering precise outcomes, future studies should prioritize the exploration of task-specific models trained for quantitative management scenarios and the use of advanced prompting techniques to enhance accuracy and reliability.

# Conclusion

This study examines the performance of LLMs in quantitative management tasks, highlighting key findings and implications for managerial decision-making support. The results show that LLMs achieve similar performance regardless of presentation format and demonstrate effective capability in handling multi-step tasks despite varying text characteristics as length. However, increasing numbers of parameters and constraints could negatively impact models' performances. Our analysis reveals clearer patterns in approximate solutions near the optimum, while exact solutions showed less consistency. While approximate answers may suffice in some managerial decision-making cases, precise answers could require tailored prompting and specially trained models to avoid performance gaps. Notably, iterative attempts did not consistently improve accuracy, reinforcing the need for careful design rather than reliance on repeated iterations. The observed performance differences between models further emphasize the importance of strategic model selection and evaluation processes. These findings underscore the need for flexible implementation strategies, particularly given rapid technological evolution and the limited range of test scenarios analyzed.

Study limitations include the constrained diversity of tested scenarios and absence of longitudinal assessments to capture performance changes over time. Future research should expand scenario diversity, incorporate domain-specific applications, and adopt longitudinal studies to better understand LLM performance trends and optimize their deployment for business decision-making.

For practitioners, this research highlights the importance of adapting LLM-based systems to quantitative task-specific complexity levels while ensuring continuous assessment as technology evolves. Organizational leaders should prioritize strategic investments in model evaluation and integration processes to maximize the effectiveness of their decision-support systems.

# References

Akter, S. N., Yu, Z., Muhamed, A., Ou, T., Bäuerle, A., Cabrera, Á. A., Dholakia, K., Xiong, C., & Neubig, G. (2023). An In-depth Look at Gemini's Language Abilities (arXiv:2312.11444). arXiv. https://doi.org/10.48550/arXiv.2312.11444

Chen, W., Yuan, C., Yuan, J., Su, Y., Qian, C., Yang, C., Xie, R., Liu, Z., & Sun, M. (2024). Beyond Natural Language: LLMs Leveraging Alternative Formats for Enhanced Reasoning and Communication (arXiv:2402.18439). arXiv. https://doi.org/10.48550/arXiv.2402.18439

Eloundou, T., Manning, M., Mishkin, P., & Clark, J. (2023). GPTs are GPTs: An early look at the labor market impact potential of large language models. Retrieved from https://doi.org/10.48550/arxiv.2303.10130

Gama, F., & Magistretti, S. (2023). Artificial intelligence in innovation management: A review of innovation capabilities and a taxonomy of AI applications. Journal of Product Innovation Management. https://doi.org/10.1111/jpim.12698

Haq, M., Ur Rehman, M. M., Derhab, M., Saeed, R., & Kalia, J. (2024). Bridging Language Gaps in Neurology Patient Education Through Large Language Models: A Comparative Analysis of ChatGPT, Gemini, and Claude. https://doi.org/10.1101/2024.09.23.24314229

He, J., Rungta, M., Koleczek, D., Sekhon, A., Wang, F. X., & Hasan, S. (2024). Does Prompt Formatting Have Any Impact on LLM Performance? (arXiv:2411.10541). arXiv. https://doi.org/10.48550/arXiv.2411.10541

Levy, M., Jacoby, A., & Goldberg, Y. (2024). Same Task, More Tokens: The Impact of Input Length on the Reasoning Performance of Large Language Models (arXiv:2402.14848). arXiv. https://doi.org/10.48550/arXiv.2402.14848

Min, B., Ross, H., Sulem, E., Pouran Ben Veyseh, A., Nguyen, T. H., Sainz, O., Agirre, E., Heintz, I., & Roth, D. (2024). Recent advances in natural language processing via large pre-trained language models: A survey. ACM Computing Surveys, 56(2), Article 30. https://doi.org/10.1145/3605943

Monea, G., Bosselut, A., Brantley, K., & Artzi, Y. (2024). LLMs Are In-Context Reinforcement Learners (arXiv:2410.05362). arXiv. https://doi.org/10.48550/arXiv.2410.05362

Nayab, S., Rossolini, G., Buttazzo, G., Manes, N., & Giacomelli, F. (2024). Concise Thoughts: Impact of Output Length on LLM Reasoning and Cost (arXiv:2407.19825). arXiv. https://doi.org/10.48550/arXiv.2407.19825

Romera-Paredes, B., Barekatain, M., Novikov, A., Balog, M., Kumar, M. P., Dupont, E., Ruiz, F. J. R., Ellenberg, J. S., Wang, P., Fawzi, O., Kohli, P., & Fawzi, A. (2024). Mathematical discoveries from program search with large language models. Nature, 625(7995), 468–475. https://doi.org/10.1038/s41586-023-06924-6

Ryu, J., Kim, J., & Kim, J. (2024). A Study on the Representativeness Heuristics Problem in Large Language Models. IEEE Access, 12, 147958–147966. https://doi.org/10.1109/ACCESS.2024.3474677

Singha, A., Cambronero, J., Gulwani, S., Le, V., & Parnin, C. (2023). Tabular Representation, Noisy Operators, and Impacts on Table Structure Understanding Tasks in LLMs (arXiv:2310.10358). arXiv. https://doi.org/10.48550/arXiv.2310.10358

Testolin, A. (2024). Can Neural Networks Do Arithmetic? A Survey on the Elementary Numerical Skills of State-of-the-Art Deep Learning Models. Applied Sciences, 14(2), 744. https://doi.org/10.3390/app14020744

Wójcik, D., Adamiak, O., Czerepak, G., Tokarczuk, O., & Szalewski, L. (2024). A comparative analysis of the performance of chatGPT4, Gemini and Claude for the Polish Medical Final Diploma Exam and Medical-Dental Verification Exam. https://doi.org/10.1101/2024.07.29.24311077

Ying, J., Lin, M., Cao, Y., Tang, W., Wang, B., Sun, Q., Huang, X., & Yan, S. (2024). LLMs-as-Instructors: Learning from Errors Toward Automating Model Improvement (arXiv:2407.00497). arXiv. https://doi.org/10.48550/arXiv.2407.00497

Yu, F., Zhang, H., Tiwari, P., & Wang, B. (2024). Natural Language Reasoning, A Survey. ACM Computing Surveys, 56(12), 1–39. https://doi.org/10.1145/3664194

Zhang, Y., Henkel, J., Floratou, A., Cahoon, J., Deep, S., & Patel, J. M. (2024). ReAcTable: Enhancing ReAct for Table Question Answering. Proceedings of the VLDB Endowment, 17(8), 1981–1994. https://doi.org/10.14778/3659437.3659452